\documentclass{article}
\usepackage{graphicx} 
\usepackage{xcolor}
\usepackage{setspace}
\usepackage{multirow}
\usepackage[normalem]{ulem}
\useunder{\uline}{\ul}{}
\usepackage{anyfontsize}
\usepackage{longtable}
\usepackage{array}
\usepackage{multirow}
\usepackage{hyperref}
\usepackage[utf8]{inputenc}
\usepackage{booktabs}
\usepackage{authblk}
\begin{document}

\title{AI-driven multi-omics integration for multi-scale predictive modeling of causal genotype-environment-phenotype relationships}
\author{%
  You Wu$^{1,*}$, Lei Xie$^{1,2,3,4,*}$
  \date{} 
  \\
  \small
  $^{1}$Ph.D. Program in Computer Science, The Graduate Center, The City University of New York, New York, New York, USA \\
  $^{2}$Ph.D. Program in Biology and Biochemistry, The Graduate Center, The City University of New York, New York, New York, USA \\
  $^{3}$Department of Computer Science, Hunter College, The City University of New York, New York, New York, USA \\
  $^{4}$Helen \& Robert Appel Alzheimer’s Disease Research Institute, Feil Family Brain \& Mind Research Institute, Weill Cornell Medicine, Cornell University, New York, New York, USA \\
  $^{*}$Correponding authors: ywu1@gradcenter.cuny.edu, lei.xie@hunter.cuny.edu  \\
}

\maketitle
\begin{abstract}
Despite the wealth of single-cell multi-omics data, it remains challenging to predict the consequences of novel genetic and chemical perturbations in the human body. It requires knowledge of molecular interactions at all biological levels, encompassing disease models and humans. Current machine learning methods primarily establish statistical correlations between genotypes and phenotypes but struggle to identify physiologically significant causal factors, limiting their predictive power.  Key challenges in predictive modeling include scarcity of labeled data, generalization across different domains, and disentangling causation from correlation. In light of recent advances in multi-omics data integration, we propose a new artificial intelligence (AI)-powered biology-inspired multi-scale modeling framework to tackle these issues. This framework will integrate multi-omics data across biological levels, organism hierarchies, and species to predict causal genotype-environment-phenotype relationships under various conditions.  AI models inspired by biology may identify novel molecular targets, biomarkers, pharmaceutical agents, and personalized medicines for presently unmet medical needs.

\end{abstract}

Keywords: machine learning, deep learning, single cell, omics data, drug discovery, precision medicine, complex disease. 

\section{Introduction}

A fundamental challenge in the field of biology is to predict phenotypes, which are the observable traits of an organism, considering the complex interaction between various genetic makeups (genotypes) and environmental influences and perturbations\cite{via1985genotype}. The genotype refers to the hereditary information stored in an organism's DNA, whereas the phenotype refers to the manifestation of that genetic information at the organismal level. In addition to genetics, environmental factors such as nutrition, pollution, infections, radiation exposure, microbiota, and drug usage play a role in shaping and altering phenotypes. 

The phenotype can be defined by observable physical characteristics (e.g., eye color), behavioral patterns (e.g., memory), physiological functions (e.g., blood pressure), and clinical manifestations (e.g., pain), among others. However, the organism's phenotype does not immediately rise from its genotype. There exist several intermediate phenotypes, known as endophenotypes\cite{gottesman2003endophenotype}, which delineate molecular attributes at an intermediate level of organization, complexity, or scale between the molecular/genetic level and the organismal phenotype. The endophenotype typically includes RNA expression, protein expression and post-translational modifications, metabolite concentrations, and so on. To establish causal linkages between genotype, environment, and phenotype, it is essential to utilize endophenotype as a means to connect the genotype and the phenotype of an organism. Firstly, the endophenotype encompasses the biological mechanisms that link the genetic makeup to the final organismal phenotype. Additionally, alterations in endophenotypes occur before or concurrently with modifications in the organismal phenotype. Therefore, endophenotypes are more responsive markers of genetic or environmental impacts compared to the organismal phenotype. Furthermore, endophenotypes are frequently able to be objectively measured and quantified. This facilitates the replication and comparison of data across many investigations, as well as the development of computer models. Ultimately, endophenotypes serve as biomarkers, such as $\beta$ -amyloid indicating Alzheimer's disease, for clinical disorders. They also offer specific targets that are linked to the causes of diseases, which in turn facilitates the development of effective and safe therapeutic interventions.

The latest development in sequencing and high-throughput technology has generated a vast amount of multiple omics data including but not limited to genomics, epigenomics, transcriptomics, proteomics, metabolomics, lipidomics, glycomics, cytomics/cellomics, microbiomics, metagenomics, radiomics, interactomics, and chemical genomics\cite{hasin2017multi}. With the exception of genomics and epigenomics data that characterize genotypes, and microbiomics, metagenomics and chemical genomics data that provide information about environmental factors, most omics data reveals the molecular landscape of distinct endophenotypes at various levels. These omics data are crucial in linking genetic information to phenotypic outcomes and predicting phenotype responses to environments. For example, analyzing transcriptomics data identifies which genes are upregulated or downregulated in response to genetic variations or environmental perturbations. Proteomics data aids in connecting genetic information to proteins that carry out most cellular functions. Metabolites are the end products of many cellular processes. They reflect the functional output of the genome, transcriptome, and proteome of an organism and environment, and capture the organism’s response to external stimuli and environmental influences. 

While each omics type provides a unique perspective on the molecular processes occurring within cells, tissues, or organisms, it is essential to combine all layers of omics data in order to fully understand the complexity and interdependencies of biological systems\cite{graw2021multi}. Firstly, rooted in the central dogma of molecular biology, it is necessary to connect multiple levels of omics data, encompassing DNAs, RNAs, proteins, and phenotypic outcomes, in order to gain a full understanding of how genetic information is converted into functional molecules and eventually, phenotypes. Secondly, integrating data across several omics levels enables identification of key regulatory elements that act as critical points of control in cellular pathways and the complex interactions and feedback loops that govern cellular processes. Finally, individual omics datasets only provide partial information about a biological system. Their integration will enhance the predictive power of computational models that aim to establish a connection between genetics and phenotype.

The human body consists of a diverse array of cell types (e.g., epithelial cells, blood cells, immune cells, etc.) The organization of the human cells follows a hierarchical structure. Cells combine to form tissues, tissues form organs, and organs work together to create a functional organism. Cells communicate through chemical signals such as hormones and neurotransmitters. The progress in single-cell and spatial omics techniques makes it possible to observe and quantify heterogeneous cellular processes and cell-cell communications across the hierarchy of an organism at a single-cell resolution\cite{cheng2023review,heumos2023best,baysoy2023technological,vandereyken2023methods}. Spatial single-cell omics data will be the cornerstone to link molecular events to organism phenotypes\cite{walsh2023decoding,bressan2023dawn,palla2022spatial,dries2021advances}. Thus, it is critical to integrate omics data across biological scales from cell to tissue to organ to organism. 
	
In addition to the integration of omics data across various biological levels and across organismal scales, it is imperative to also integrate omics data across different species\cite{song2023benchmarking,yang2023genecompass,rosen2024toward}. Omics studies conducted in model systems are essential for enhancing our understanding of biology. This is due to specific advantages offered by model organisms, such as short generation times, ease of genetic manipulation, and similarities to more complex organisms including humans. Model organisms have long been instrumental in investigating the functions of specific genes and the regulatory mechanisms that control them. They have also helped in comprehending cellular processes, tissue formation, and organ development, as well as shedding light on the genetic factors influencing complex behaviors. Genetically engineered models are indispensable tools for understanding the molecular mechanisms underlying diseases, evaluating possible treatments, and assessing the safety and efficacy of therapeutic interventions. As stated by Theodosius Dobzhansky, “Nothing in biology makes sense except in the light of evolution”\cite{dobzhansky2013nothing}. Comparative genomics studies identify the similarities and differences in the genomes of different species, leading to understanding the genetic basis of human traits and diseases. Recent advances in functional genomics such as CRISPR-Cas9 and perturb-seq make it possible to assess gene functions and dissect gene regulatory networks on a large scale. As the amount of multi-omics data from model organisms becomes more accessible, there is a need for innovative methods to transfer this knowledge from model organisms to humans. This will facilitate advancing fundamental and translational biomedical sciences.

Cross-level, cross-scale, cross-species multi-omics data integration and predictive modeling of causal genotype-environment-phenotype relationships will not only provide new knowledge about the basic principles of life but also be the driving force for the identification of new molecular targets, biomarkers, pharmaceutical agents, and personalized medicines for unmet medical needs, as shown in Figure 1. The target-based drug discovery and development approach, which follows the human genome revolution and now dominates the pharmaceutical industry, is widely recognized for its time-consuming, expensive, and unproductive nature. Artificial Intelligence (AI) shows great potential in expediting the process of drug discovery. Nonetheless, if adhering to the current drug discovery paradigm, AI may merely make failures faster and cheaper but not improve the success rate of identifying effective and safe treatments for incurable ailments. As highlighted by Bender et al.\cite{bender2021artificial}, there exists a substantial disparity between molecules that are optimized for target binding affinities or other proxy objectives and medications that demonstrate both clinical efficacy and safety. A survey conducted recently indicates that over 90\% of the medications that have been approved are derived from the process of phenotype drug discovery and development\cite{moffat2017opportunities}. However, conventional phenotype screening approaches based on cellular or organismal phenotypes are medium- or low-throughput and lack information about drug modes of action. 

Perturbation functional omics profiling offers a quantitative, mechanistic, and high-throughput phenotype readout for compound screening, thereby boosting the power of phenotype drug discovery\cite{vincent2022phenotypic}. Nevertheless, the molecular omics profiles only characterize endophenotypes. It remains a challenge to connect endophenotypes to clinical outcomes. Moreover, due to the impracticality of measuring compound responses in a living patient, we must rely on a model system during the early and pre-clinical stages of drug discovery and development. In order to accurately translate chemical activity observed in a model system to the effectiveness and side effects of a medicine in a clinical setting, it is necessary to have an unbiased and comprehensive phenotype readout that bears information of target transferability, drug mode of action and pharmacokinetics. An integrated molecular profile derived from transcriptomic, proteomic, metabolomic, and other endophenotypes both before and after the treatment of tested compounds is highly suitable for this objective and has shown potential in phenotype compound screening\cite{deepce_pham2021deep} and physiologically based pharmacokinetics\cite{chen2022integration}.

\begin{figure}
   \hspace{-3cm}
    \includegraphics[width=1.5\textwidth]{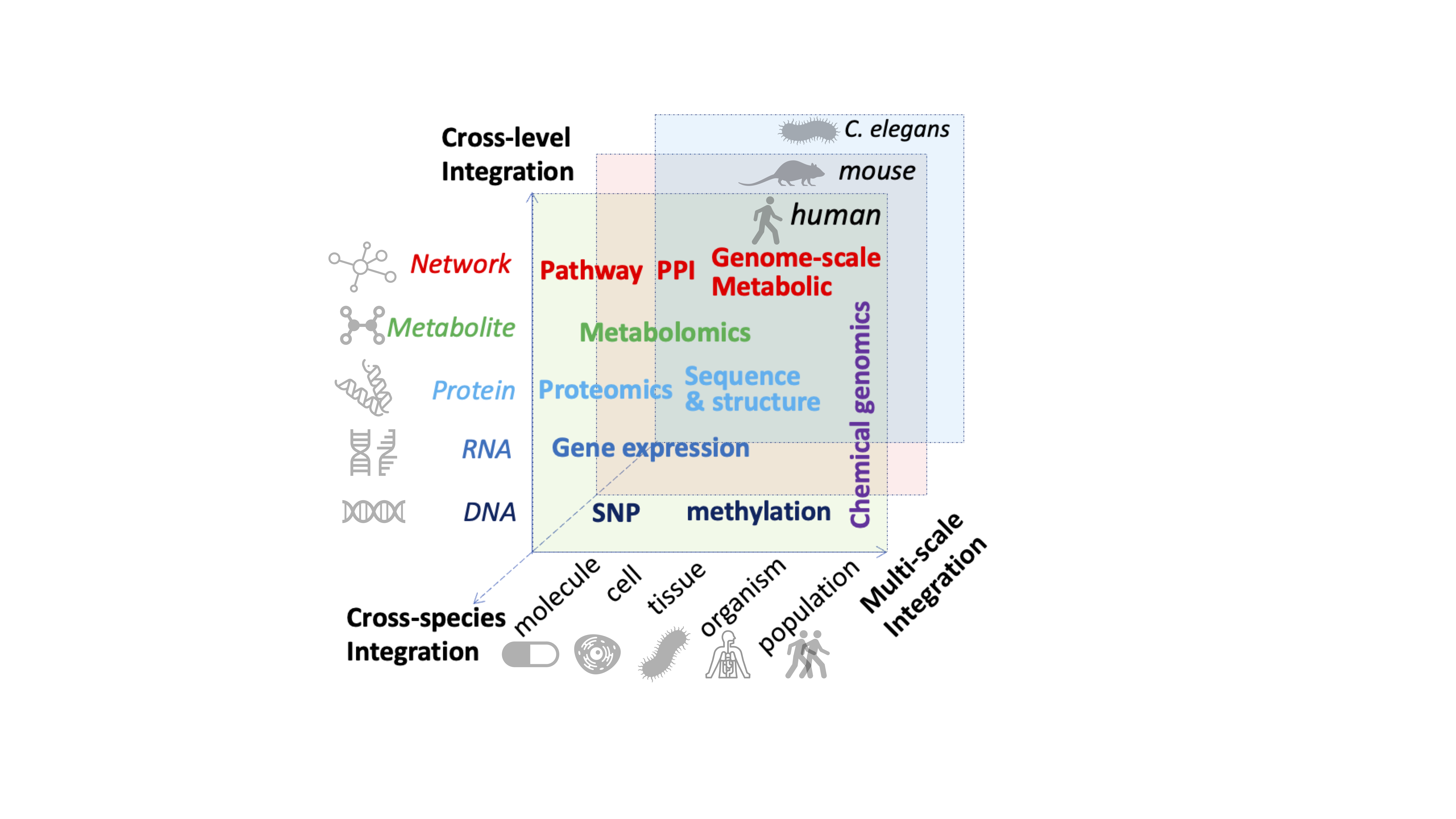}
    \caption{Illustration of cross-level, cross-scale, cross-species multi-omics data integration}
    \label{fig:integration}
\end{figure}

In summary, elucidating genetic and molecular underpins of complex human traits and disorders, as well as predicting organismal phenotypes under the interplay of diverse genotypes and environmental perturbations, requires integrating multiple omics data across data modalities, biological levels, and species \ref{fig:integration}. In this review paper, we first summarize available perturbation omics data and examine recent advances in machine learning techniques for integrating and analyzing multi-omics data. Then we shed light on limitations inherent in current methodologies. Finally, we propose a new biology-inspired Artificial intelligence (AI)-driven framework for multi-omics integration and multi-scale predictive modeling, aimed at predicting human phenotypic responses to unprecedented perturbations. The framework holds promise in illuminating fundamental principles of life and discovering new molecular targets, biomarkers, pharmacological agents, and personalized therapies for presently untreatable diseases.

\section{Perturbation omics data resources}
Predictive modeling of phenotypes from genotypes under perturbations needs labeled data. A large number of perturbation omics data have been collected. Although these data are highly biased to certain biological conditions (cell types, diseases, etc.) and perturbation types, they are the starting point for machine learning. Several representative data sets are listed in Table \ref{tab:data} and summarized below.

TCGA (The Cancer Genome Atlas) \cite{TCGA_tomczak2015review} is a comprehensive resource that has molecularly characterized thousands of primary cancers and matched normal samples across numerous cancer types. By integrating data on genetic mutations, gene expression, methylation, and protein profiles, TCGA provides a robust framework for understanding the molecular underpinnings of cancer, aiding in identifying biomarkers and therapeutic targets.

LINCS (Library of Integrated Network-based Cellular Signatures) within the Connectivity Map (CMAP) project aims to elucidate cellular responses to a variety of perturbations, such as small molecule treatments and genetic modifications. By employing high-throughput techniques, LINCS generates vast datasets on gene expression and protein levels, LINCS Data Portal provides access to LINCS data from various sources \cite{lincs_portal1_koleti2018data,lincs_portal2_stathias2020lincs}.

DepMap (Dependency Map) \cite{DepMap_tsherniak2017defining}is a pioneering project that systematically identifies the genetic and chemical dependencies of cancer cells. Through high-throughput CRISPR-Cas9, RNAi, and chemical screens, it maps essential genes and pathways critical for cancer survival. This comprehensive resource integrates data from various platforms, including the Cancer Cell Line Encyclopedia (CCLE)\cite{ccle1_barretina2012cancer,ccle2_ghandi2019next}, Genomics of Drug Sensitivity in Cancer (GDSC)\cite{gdsc1_garnett2012systematic,gdsc2_iorio2016landscape}, and the Cancer Therapeutics Response Portal (CTRP)\cite{ctrp2_basu2013interactive, ctrp1_seashore2015harnessing}.

scPerturb\cite{peidli2024scperturb} is dedicated to single-cell perturbation studies, offering detailed insights into how individual cells respond to genetic modifications and other perturbations. By utilizing advanced single-cell RNA sequencing techniques, scPerturb captures the heterogeneity and dynamic responses of single cells, helping researchers decipher gene function, regulatory networks, and the impact of genetic changes at an unprecedented resolution. ScPerturb offers an integrative dataset from 44 published works including various methods and Sci-Plex\cite{sciplex_srivatsan2020massively}.

PharmacoDB\cite{smirnov2018pharmacodb} is an integrative database that consolidates pharmacogenomic data from multiple high-throughput drug screening studies. It provides a platform for exploring drug responses across various cancer cell lines, facilitating the identification of drug efficacy, resistance mechanisms, and potential biomarkers. PharmacoDB supports personalized medicine by enabling researchers to connect molecular profiles with drug sensitivity data, promoting the development of tailored therapeutic strategies.

ProteomicsDB\cite{lautenbacher2022proteomicsdb1} is a comprehensive database dedicated to the integration of human proteomic data. It consolidates data from multiple high-throughput proteomics experiments to provide detailed information on protein expression, post-translational modifications (PTMs), and protein interactions. The platform also features recent studies on decrypting the molecular basis of cellular drug phenotypes (DecryptE\cite{eckert2024decryptE}), as well as drug actions and protein modifications by dose- and time-resolved proteomics (DecryptM\cite{zecha2023decryptM}).

\newpage
\begingroup 
\fontsize{9pt}{11pt}\selectfont   
\setlength\LTleft{-3.5cm}
\begin{longtable}{p{3cm} p{3cm} p{4cm} p{4cm} p{4cm}}
\hline
\textbf{Source} & \textbf{Perturbation Type} & \textbf{Molecular \ Profiling}&\textbf{Assay Readout} & \textbf{Datasets Included} \\ \hline
\textbf{TCGA\cite{TCGA_tomczak2015review}} & Drug & Genomic, transcriptomic, epigenomic, proteomic & Clinical and survival data & 33 tissue types \\
\textbf{LINCS Data Portal\cite{lincs_portal1_koleti2018data,lincs_portal2_stathias2020lincs}} & Drug, CRISPR-Cas9, ShRNA & Perturbed transcriptomic, proteomic & Transcriptomic, proteomic, kinase binding, cell viability, cell growth inhibition, apoptosis, morphology & LINCS 1000, LINCS proteomic, ChEMBL\textsuperscript{*}, Tox21\textsuperscript{*}, Cell Painting morphological profiling assay\textsuperscript{*}\\
\textbf{DepMap\cite{DepMap_tsherniak2017defining}} & CRISPR-Cas9, RNAi screen, drug & Genomic, transcriptomic, proteomic & Perturbed genomic, transcriptomic, proteomic, drug sensitivity, drug response  & CCLE, GDSC, CTRP \\
\textbf{scPerturb\cite{peidli2024scperturb}} & CRISPR-cas9, CRISPRi, CRISPRa, TCR stim, cytokines, drug & Single-cell RNA sequencing (scRNA-seq), proteomic, epigenomic & Perturbed scRNA-seq, proteomic, chromatin accessibility & Sci-plex, 44 public single-cell perturbation datasets \\
\textbf{PharmacoDB\cite{smirnov2018pharmacodb}} & Drug & Genomic, transcriptomic, proteomic  & Drug sensitivity, drug response & CCLE, GDSC, NCI-60, PRISM, FIMM, GTRP, GRAY, gCSI \\ 
\textbf{ProteomicsDB\cite{lautenbacher2022proteomicsdb1}} & Drug & Proteomics, transcriptomics & Posttranslational modifications (PTMs), perturbed proteomics, phenomics &  DecryptE, DecryptM, GeneCards\textsuperscript{*}, UniProt\textsuperscript{*}, OmniPathDB\textsuperscript{*} and Gene Information eXtension (GIX)\textsuperscript{*}
\\ \hline
\caption{Perturbation data resource, *linked data resource}
\label{tab:data}
\end{longtable}
\endgroup

\newpage
\begingroup 
\fontsize{9pt}{7pt}\selectfont   
\setlength\LTleft{-3.5cm}

\begin{longtable}{p{2cm}|p{2.5cm}p{2.5cm}p{3cm}p{6cm}} \hline



Learning &Methods &Representative Papers& Modality& Notes \\ \hline

\multirow{12}{2cm}[-7cm]{ Unsupervised} & \multirow{7}{2cm}[-3.3cm]{Autoencoder}  & ScVI\cite{scvi_lopez2018deep}& scRNA-seq & Effective in removing batch effects; however, it is constrained to analyzing only single modality data                                                                                                 \\
                                                         &                                   & scANVI\cite{scanvi_sxu2021probabilistic}                          & scRNA-seq                                                                                                 & Facilitates label transfer with uncertainty measures in semi-supervised learning; limited to a single modality       \\
                                                         &                                   & TotalVI\cite{totalvi_gayoso2021joint}                         & scRNA-seq, surface protein                                                                   & Learns a joint probabilistic representation of both RNA and proteins; but requires paired measurements and does not align domains across different experiments  \\
                                                         &                                   & Cobolt\cite{gong2021cobolt}                          & mRNA-seq, scRNA-seq, ATAC-seq, scATAC-seq                                                                           & Offers guided multimodal integration for paired RNA-seq and ATAC-seq data, but the assumptions of a multinomial distribution might ignore the biological context of different modalities                                     \\
                                                         &                                   & MultiVI\cite{ashuach2023multivi}                         & scRNA-seq, scATAC-seq, surface protein                                          & Guides multimodal integration accounting for modality-specific noise; uses a symmetric approach for joint representation, though affected by data sparsity                            \\
                                                         &                                   & scMVP\cite{scmvp_li2022deep}                           & scRNA-seq, scATAC-seq                                                                        & Provides non-symmetric multimodal integration with multi-head attention and cycle-GAN; but requires paired sample data                                                                                        \\
                                                         &                                   & GLUE\cite{glue_cao2022multi}                            & scRNA-seq, scATAC-seq                                                                           &  Triple-omics integration while simultaneously inferring regulatory interactions; adversarial training may lack stability                                                            \\
                                                         &                                   & Biolord\cite{biolord_piran2024disentanglement}          &  scRNA-seq, drug, dosage, cell line                                                            & Encodes cellular identity attributes separately for better representation; needs exploration of unknown attributes to improve generalizability                                                            \\                                   &                                   & ChemCPA\cite{chemcpa_hetzel2022predicting}                    & Bulk \& scRNA-seq, drug, dosage                                                                          & Incorporates compound structure and bulk RNA-seq data with adversarial training to adapt to single-cell data; effective for unseen compounds but needs evaluation on unseen cell lines                                                            \\     \cline{2-5}                                                    
                                                         & \multirow{2}{*}{Transformer}      & scGPT\cite{cui2024scgpt}                           & scRNA-seq, scATAC-seq, surface protein, Perturb-seq                                               & Foundation model trained on over 10M cells, capable of learning cell-specific information; requires paired data and limited reliability in zero-shot settings                                        \\
                                                         &                                   & GeneCompass\cite{yang2023genecompass}                     & Cross-species, scRNA-seq, perturb-seq, LINCS1000                                                         & Foundation model trained over 12OM cells cross-species incorporating prior knowledge; confined to transcriptomic data                                  \\
                                                         & \multirow{3}{2cm}[-1.5cm]{Other techniques} & SATURN\cite{saturn_rosen2024toward}                          & Cross-species, scRNA-seq, protein sequence                                                                & Enables cross-species analysis by merging protein language models with scRNA data; challenges exist due to the absence of direct orthologs and it requires paired data                    \\ \cline{2-5}
                                                         &                                   & scCLIP\cite{xiong2023scclip}                          & scRNA-seq, scATAC-seq                                                                                      & Employs contrastive learning for multimodal single-cell data; paired sample data is mandatory                                                                                                    \\
                                                         &                                   & MatchCLOT\cite{gossi2022matchclot}                       & scRNA-seq, scATAC-seq, surface protein abundance                                                          & Combines contrastive learning with optimal transport; reliant on paired sample data                                                                                                              \\ \hline
\multirow{9}{*}{Supervised} & \multirow{5}{*}{Multimodal}       & Yang et al.\cite{yang2021multi}                     & Image, RNA-seq, ATAC-seq, Hi-c                                                                           & Integrates various data types for cancer models; each model is specific to one type of cancer and requires paired data                                                                                                                                      \\
                                     &                                   & Faisal et al.\cite{faisal_chen2022pan}                  & H\&E WSIs and molecular profile features                                                                  & Correlates histopathological images with molecular profiles; demands paired data and is specific to individual cancer models                                                                                                                                      \\
                                     &                                   & DSIR\cite{DSIR_yang2022deep}                            & DNA methylation, mRNA and miRNA expression                                                                & Utilizes a similarity matrix for cancer subtyping; dependent on paired data and tailored to individual cancers                                                                                   \\
                                   &                                   & DLSF\cite{DLSF_zhang2022deep}                            & DNA methylation, mRNA and miRNA expression                                                                & Applies a cycle autoencoder to extract a consistent sample manifold at the multi-omics level; also requires paired data for each cancer model                              \\
                                   &                                   & MOMA\cite{moon2022moma}                            & DNA methylation, mRNA and miRNA expression                                                                            & Processes genes and methylation data using a geometric approach; models need to be individually trained for each cancer type and paired data is needed                                                          \\ \cline{2-5}
                                     & \multirow{4}{2cm}[-2cm]{Knowledge graph and other techniques} & Lee et al.\cite{cellcell_lee2024cell} & Bulk \& scRNA-seq                                                                                   & Develops patient-specific cell-cell communication networks to predict immune checkpoint inhibitors efficacy and uncover key pathways; yet, it simplifies complex network relationships                    \\
                                     &                                   & BioBridge\cite{wang2023biobridge}                       & Protein, molecule, disease, biological process, molecular function, and cellular component & Leverages knowledge graphs to transition between unimodal foundations without fine-tuning; lacks quantitative evidence for molecular generation tasks \\
                                    &                                   & One for all\cite{ofa_liu2023one}                     & Literature category description, molecule property description, relation type description                 & Constructs text-attributed graphs for diverse cross-domain associations; it does not meet the state-of-the-art performance for individual tasks                                                                  \\ 
                                    &                                   & GEARS\cite{gears_roohani2023predicting}                     & Gene-gene interaction, scRNA-seq                 & Integrates GNN with a gene-gene interaction knowledge graph; limited to the same cell type and experimental condition, with confounding factors from combinatorial perturbational data         \\\hline
\caption{Representative state-of-the-art computational methods for multi-omics data integration toward predictive modeling of genotype-environment-phenotype relationships}
\label{tab:sota}
\end{longtable}
\endgroup

\newpage

\section{State-of-the-art of machine learning methods for multi-omics data integration and predictive modeling}

\subsection{Unsupervised learning}
One of the major technical challenges faced by multi-omics data integration is data distribution shifts. The data shift in omics data mainly comes from two sources: technical confounders such as batch effects and biological confounders (e.g., sex, age, disease state, etc.).  Traditional statistical methods provided a foundation for multi-omics data integration. These approaches encompass a variety of techniques including correlation-based analysis (e.g., BindSC\cite{dou2022bindsc}, Seurat v3\cite{stuart2019seuratv3}, Scanorama\cite{scanorama_hie2019efficient} and MaxFuse\cite{chen2023maxfuse}), matrix factorization (iNMF\cite{kriebel2022uinmf_liger} and LIGER\cite{liger_liu2020jointly}), Bayesian-based methods (MOFA+\cite{mofa_argelaguet2020mofa+}), nearest neighbor-based (e.g., fastMNN\cite{fastmmn_haghverdi2018batch} and Seurat v4\cite{seuratv4_hao2021integrated}), and dictionary learning (e.g., Seurat v5\cite{seuratv5_hao2023dictionary}).  Our focus, however, is on deep representation learning methods, which have shown great promise in addressing the aforementioned challenges. The representative techniques include autoencoder, transformer, and contrastive learning. The power of these methods comes from the fact that they do not need labeled phenotypic data that is scarce, and often infeasible to obtain. 

\subsubsection{Autoencoder}
Deep generative models, particularly Variational Autoencoders (VAEs), are at the forefront of analyzing complex, high-dimensional single-cell sequencing data. VAEs employ an encoder to interpret input data and a decoder to reconstruct it, learning a latent distribution. The objective that it optimizes is to mirror the input while minimizing the Kullback-Leibler divergence between the latent embedding's prior and posterior distributions.

scVI\cite{scvi_lopez2018deep} models gene expression in scRNA-seq data using VAE with a zero-inflated negative binomial distribution, conditioned on batch annotations and two unobserved variables: a cell-specific scaling factor and a latent biological variable. Neural networks map these latent variables to the distribution parameters, producing batch-corrected, normalized transcript estimates for differential expression and imputation. A separate neural network, trained via variational inference and stochastic optimization, approximates the posterior distribution of latent variables, ensuring scalable and accurate analysis of single-cell RNA-seq data.

The same group further developed scANVI\cite{scanvi_sxu2021probabilistic}, which integrates semi-supervised learning with cell type annotations. It can be useful for transfer labels while measuring uncertainty, especially when dealing with complex label structures such as hierarchical cell types. However, both models are limited on RNA-seq data as a single modality.

TotalVI\cite{totalvi_gayoso2021joint} took advantage of the CITE-seq technique, which can simultaneously measure the abundance of the proteins on the cell surface,  to provide the opportunity for multifaceted analysis of both RNA-seq and the functional information in proteins. It uses VAEs to learn a joint probabilistic representation of the paired measurements that counts for batch effects for both modalities. The RNA modeling strategy is similar to scVI \cite{scvi_lopez2018deep}. The protein modeling explicitly has modality-specific technical factors such as a protein background, which enable a denoised view of data. However this method requires paired measured samples, nor there is domain alignment consideration.

More recent tool Cobolt\cite{gong2021cobolt} introduces a symmetric multi-modal VAE network for multi-omics data integration with a Product of Experts model (PoE) model\cite{poe_hinton2002training}. PoE combines the variational posteriors of the multiple modalities (the experts) by taking their product and normalizing the result. It was trained on paired multi-omics data to guide the integration of unpaired data, resulting in a joint representation of single-cell RNA-seq and ATAC-seq datasets, which can be beneficial for various downstream tasks.  Despite its guidance on the unpaired datasets, this method assumed a multinomial distribution for both modalities which may cause potential information loss.

In contrast, MultiVI\cite{ashuach2023multivi} employs a modality-specific noise system suited to both gene expression and chromatin accessibility, with negative binomial distribution and Bernoulli distribution respectively.  In contrast to Cobolt's PoE technique, MultiVI utilizes a distributional mean and penalization strategy for a more optimized integration of latent embeddings. Moreover, its ability to incorporate cell surface protein abundance broadens its scope, allowing for a richer understanding of cellular properties.

The strengths of both MultiVI and Cobolt, which implemented symmetric multimodal VAE for joint modality representations, are tempered by the challenges of extreme sparsity and random noise in the datasets. These factors can confound the biological variance, posing obstacles to downstream analysis and scalability of the model. Addressing this, scMVP\cite{scmvp_li2022deep}, employs a non-symmetric framework that enables the construction of a unified latent space for scRNA-seq and scATAC-seq data. This is achieved via a clustering consistency-enforced multi-view VAE, which is further enhanced by multi-head self-attention mechanisms and a cycle-GAN module, thereby increasing the robustness across both modalities.  However, it again requires simultaneous multi-modality measurements with individual cells to function effectively.

To address the challenge of information loss when integrating data across different modalities, GLUE\cite{glue_cao2022multi} employs a modality-specific graph VAE to refine the feature transformation process by modeling regulatory relationships between chromatin regions and genes. It learns not only local but also global information. With a scalable adversarial alignment, GLUE also enables the integration of three modalities such as gene expression, chromatin accessibility, and DNA methylations. 

Biolord\cite{biolord_piran2024disentanglement} is a deep generative method designed to predict cellular responses to unseen drugs and genetic perturbations. It uses an autoencoder to separately encode multiple attributes of cellular identity, along with a single encoding for unknown attributes. This setup defines a decomposed latent space, serving as the input for the generative module to provide measurement predictions. The authors claim this design disentangles the representation with respect to known attributes. However, further exploration of the representation of unknown attributes would enhance the model's generalizability.

Hetzel et al. introduced ChemCPA\cite{chemcpa_hetzel2022predicting} a model that incorporates knowledge about compound structures and transfers bulk RNA-seq data into both identical and different gene sets between source (bulk) and target (single-cell) datasets. It uses an encoder-decoder architecture with adversarial training, allowing the model to disentangle representations of various attributes and study the effects on specific sources. Although the model was evaluated on unseen compounds, it would be more interesting if it could also work on unseen cell lines.

\subsubsection{Transformer}

In research areas such as natural language processing (NLP) and computer vision (CV), Transformer as highlighted by the attention mechanism has gained significant attention in recent years, as evidenced by its successful deployment in foundation models. Pioneering models such as BERT\cite{devlin2018bert}, GPT\cite{gpt3_brown2020language,achiam2023gpt4}, PaLM\cite{anil2023palm,chowdhery2023palm}, and LLaMA\cite{touvron2023llama} have set benchmarks in NLP as well as DALL-E\cite{dalle_pmlr-v139-ramesh21a} in CV have made significant contributions to various downstream tasks. In a biological context,  similar to how words construct a sentence, genes construct cells. Analogous to how natural language acts as a foundational layer for interpreting human behavior, the transcriptome similarly serves as a fundamental layer for unraveling the intricacies of gene regulatory mechanisms in biology. Studies have utilized single-cell transcriptomic data to construct pre-trained foundation models, such as scGPT\cite{cui2024scgpt}, Genefomer\cite{geneformer_theodoris2023transfer} and scFoundation\cite{scfoundation_hao2023large}. The representative work scGPT constructed the first foundation model through pretraining on over 10 million cells with a 12-layered transformer architecture. It also supports multiple omics data integration from paired data sources. The utilization of the self-attention approach over genes enables the encoding of gene-gene interaction, and the cell conditional tokenization also allows the model to learn cell-specific information, such as different batches and sequencing modalities. However, this technique is constrained by paired data, and exhibits limited reliability in zero-shot settings \cite{kedzierska2023assessing}.
 
While foundation models have achieved notable successes in a variety of downstream tasks, their potential has not yet been leveraged for cross-species data integration. However, the conserved nature of gene regulatory mechanisms across different species presents an outstanding opportunity to delve into the complexities of gene regulation through such integrative analysis. Bridging the cross-species analytical gap, GeneCompass\cite{yang2023genecompass} emerges as an innovative foundation model, extensively pre-trained on a vast dataset comprising over 120 million single-cell transcriptomes from human and mouse origins. It integrates gene IDs, expression values, and prior knowledge together as gene tokens, implementing a 12-layer transformer model for encoding encoding. It also facilitates a variety of downstream tasks through supervised learning, encompassing gene regulatory network elucidation, predictions of drug effects, gene dosage implications, and cellular responses to perturbations. However, GeneCompass is limited to transcriptomics data.

\subsubsection{Other techniques (contrastive learning etc.)}
SATURN\cite{saturn_rosen2024toward} stands out as the first model that combines protein embeddings, generated using large protein language model ESM2\cite{esm2_lin2023evolutionary}, with gene expression from scRNA-seq datasets. Overcoming the challenges of absent direct one-to-one orthologs, it couples protein embeddings with gene expression, employing soft clustering to form ‘macrogene’ groups. This approach allows the model to learn universal cell embeddings that bridge differences between individual single-cell experiments even when they have different genes. It combines training with conditional autoencoders with ZINB loss inspired by Lopez et al.\cite{scvi_lopez2018deep}, and other learning metrics by forcing the different cells within the same dataset far apart using weakly supervised learning and similar cells across the dataset closer to each other in an unsupervised manner. But it requires paired information.
 
scCLIP\cite{xiong2023scclip} introduces a novel application of transformers to scATAC-seq data, drawing inspiration from the contrastive learning principles of CLIP\cite{clip_radford2021learning}, it trains a pair of transformer-based encoders on multimodal single-cell data, utilizing a contrastive loss function for optimization. The result is scCLIP’s adeptness at integrating multimodal data into a singular, unified embedding space, with the scalability to accommodate extensive tissue and organismal data from large-scale atlas projects. 

Recent applications of optimal transport (OT) in single-cell data analysis have enabled the identification of cellular dynamics and the alignment of multi-omics datasets. MatchCLOT\cite{gossi2022matchclot} leverages these advancements by training two modality-specific encoders to project single-cell multimodal measurements onto a unified latent space. A novel OT algorithm is then employed for the soft-matching of cells between modalities, using batch labels to narrow the search space and mitigate distribution shifts. 

\subsection{Supervised learning}
\subsubsection{Multi-modal supervised learning}

Yang et al.\cite{yang2021multi} propose a method using autoencoder across different modalities to achieve integration, each modality is encoded using a local network, such as a convolutional network for image data, fully connected network for sequence data (RNA-seq and ATAC-seq), graph convolutional network for Hi-C. The joint representations are learned from the shared latent space, facilitating the translation between different modalities via a combination of encoders and decoders. 

Faisal et al.\cite{faisal_chen2022pan} adopt a deep learning-based multimodal fusion algorithm to integrate H\&E whole slide images (WSIs) and molecular profile features, including Copy-Number of Variation (CNV), RNA-seq, and Mutation Status (MUT). Their method is particularly rigorous for its comprehensive application in survival prediction and patient risk stratification, enhanced by a focus on interpretability through the analysis of feature importance and gene attributions. 

Deep Subspace Integration Representation (DSIR)\cite{DSIR_yang2022deep} represents another technique for multi-modality integration, utilizing deep subspace learning to simultaneously learn the local and global structures. By constructing a consensus similarity matrix, DSIR finetunes its model for cancer subtype identification through spectral clustering. 

Similarly, DLSF\cite{DLSF_zhang2022deep}  also obtains the self-representation coefficient matrix for disease subtype identification, what it differs from DSIR is the exploration of the shared global similarity structure, because DLSF uses cycle autoencoders with a shared self-expressive layer to adaptively extract a consistent sample manifold a multi-omics level. 

Moreover, A geometrical approach Module-based Omics Data Integration MOMA\cite{moon2022moma} vectorizes genes and modules, using the vector sum of genes within a module to represent it. The incorporation of an attention mechanism as a mediator allows the model to identify the most related modules among multiple omics data types, by training with various tasks of predicting phenotypes. 

For all the multi-modal techniques mentioned above, despite their potential for cross-modal integration, those approaches require paired data from the various modalities and are tailored to individual cancer types, limiting their generalizability. 


\subsubsection{Knowledge graph and other techniques}

Graph (network) representation has been widely applied in systems biology to represent biological organizations and interactions\cite{Marinka_li2022graph}. It is successful in integrating diverse types of biological and chemical data for representing genotype-environment-phenotype relationships\cite{lei_lee2020heterogeneous}. Compared with multi-modal supervised learning, graph learning directly encodes complex interactions between entities and captures semantic relationships underlying data. This allows for the seamless integration of information from diverse sources, the deduction of new information based on existing knowledge, and a deeper understanding of context and interconnections between entities. 

Lee et al.\cite{cellcell_lee2024cell} propose a machine learning model to predict cancer response to immune checkpoint inhibitors (ICIs). The network is constructed on cell-cell communication with cell types as nodes and communication strength as edges, which is deconvoluted from the patient’s bulk tumor transcriptomics data. The model can also identify key communication pathways that are consistent with single-cell level information. However, the graph is shallowly designed and more sophisticated deep learning models could be utilized to reveal complex relationships.

BioBridge\cite{wang2023biobridge} is representative of the integration of multimodal foundation models. To overcome the singularity of foundation models by applying knowledge graphs to learn the transformation between one unimodal foundation model and another, and only the bridge module needs training while all the base foundation models are kept fixed, resulting in great computational efficiency.  A various range of prediction tasks can be performed via BioBridge including cross-modality retrieval tasks, semantic similarity inference, protein-protein interaction, and cross-species protein-phenotype matching. But it lacks quantitative evidence for molecular generation tasks.
		 	 	 		
The OFA\cite{ofa_liu2023one}approach suggests using text-attributed graphs to represent the diverse cross-domain attributes and connections in a graph to combine various types of graph data. This method involves converting these descriptions into feature vectors in the same embedding space using language models, regardless of their original domain. Additionally, the method introduces "nodes-of-interest" to standardize how we approach different graph-related tasks using a single task. OFA also uses a unique method called graph prompting by adding special structures to the graph that act like prompts, allowing the model to perform a wide range of tasks without fine-tuning. The model is designed to handle various fields, such as citation networks, molecular structures, and knowledge bases. Despite the strengths of this method, the performance for individual tasks seems suboptimal.

Integrating deep learning with a knowledge graph of gene-gene interactions, GEARS\cite{gears_roohani2023predicting} predicts transcriptional responses to both single and multigene perturbations using single-cell RNA sequencing data from perturbational screens. It employs a Graph Neural Network (GNN) to study genetic relationships and perturbational expression changes, enabling predictions for gene combinations not experimentally perturbed. However, the model is limited to the same cell type or experimental condition, and its reliance on combinatorial perturbational data introduces confounding factors that need further addressing.

\section{Challenges in machine learning techniques}

Despite significant progress in applying machine learning to the integration of multi-omics data and predictive modeling of genotype-environment-phenotype relationships, several challenges persist. These include the need for biologically informed representation learning, scarcity and ambiguity of labeled data, inability to generalize out-of-distribution, and dealing with incomplete and noisy graphs. 

\subsection{Need for biologically informed representation learning}
A fundamental hurdle arises from the multi-level hierarchical organization of biological systems, as discussed in the Introduction section. On one hand, multiple statistically insignificant variations at a lower level can collectively result in significant changes at a higher level (e.g., gene expression)\cite{Polygenic_article}. Hence, a network biology approach is imperative to enhance biological signals\cite{cowen2017network}. On the other hand, many genotypes exert a pleiotropic effect on complex diseases and traits\cite{sivakumaran2011abundant}. Consequently, a higher-level endophenotype demonstrates greater discriminatory power concerning the organismal phenotype than a lower-level one. Therefore, a cross-level modeling approach is necessary to simulate the asymmetrical information transmission process between genotype and phenotype\cite{he2022cross}. This, in turn, will enhance model interpretability and facilitate the elucidation of molecular underpinnings of phenotypes\cite{yang2019white,liu2021transynergy}.


\subsection{Scarcity and ambiguity of labeled data}
The scarcity of labeled data significantly hinders the application of machine learning in the predictive modeling of genotype-environment-phenotype relationships through multi-omics data. Current multi-modal learning often necessitates paired omics data with shared labels, a challenge exacerbated by the infrequent availability of such labeled data in many instances. For example, transcriptomics and proteomics data from the brain tissues of Alzheimer’s disease patients can only be obtained from post-mortem persons. Consequently, constructing a practical machine learning model for living patients relies on genomics or brain imaging data, despite transcriptomics and proteomics data exhibiting stronger predictive power for phenotypic responses to drug treatments and other environmental influences than genomics and brain imaging data.

The issue of phenotype label ambiguity is a concern that has not received sufficient attention in machine learning. Recent efforts, including the Phenotype and Trait Ontology (PATO)\cite{gkoutos2018anatomy}, pave the way to address this problem. PATO provides a standardized vocabulary for describing phenotypic qualities in a manner that can be consistently applied across different species. However, additional efforts are needed to incorporate ontologies into machine learning models. 

\subsection{Inability to generalize out-of-distribution}
A more pressing data issue emerges with an out-of-distribution (OOD) scenario, where new unseen cases differ significantly from the data used to train the model\cite{liu2021towards}. Technological limitations and human biases have illuminated only a fraction of the vast biological and chemical universe. For instance, among over 20,000 human genes, only proteins encoded by hundreds of genes have known small molecule ligands, without accounting for isoforms, protein complexes, mutation states, and conformations. Despite an estimated $10^{60}$ small organic molecules in the chemical space, only approximately $10^8$ have known bioactivities. Single-cell profiling techniques have generated omics data for numerous cell types, but only around 100 of them have controlled perturbations and functional genomics readouts. The combined space of chemicals, biomolecules, and endo- and organismal phenotypes is staggeringly vast \cite{lunke2023integrated}. 

Another significant issue arises due to a notable distribution shift from \textit{in vitro} to \textit{in vivo} settings. This shift often results in disease models failing to accurately reflect the efficacy and toxicity of drugs in humans. There is a critical need for a computational approach that can effectively disentangle confounding factors while preserving unique features. Existing methods that fail to adequately address confounding factors often overlook their connection to clinical outcomes. A more systematic approach is required to address this challenge. 

To address the OOD problem, it becomes imperative to quantify the prediction uncertainty of new cases\cite{gawlikowski2023survey,seoni2023application}. Uncertainty quantification is particularly critical in high-stakes applications like drug discovery and precision medicine. Given the resource-intensive nature of drug discovery, uncertainty quantification aids in decision-making by offering insights into the confidence levels associated with predictions. In precision medicine, where erroneous predictions about drug efficacy or safety can have severe consequences, uncertainty quantification is essential for assessing the risks associated with model predictions.

\subsection{Incomplete and noisy graphs}
In the realm of predictive modeling for genotype-environment-phenotype relationships, two key issues within graph learning remain inadequately addressed: the incorporation of novel nodes lacking previously recognized connections in an established graph model and the identification of dubious or conflicting relationships. 

The construction of a high-quality graph model for a biological system is a labor-intensive, domain-specific task that often demands manual data curation. Furthermore, the graph model may fall short in capturing implicit knowledge and intricate patterns not explicitly represented in the data, restricting its ability to unveil novel discoveries. This limitation is particularly critical in biology, where a vast number of biological and chemical entities remain uncharted, lacking any annotations. These unannotated nodes become isolated in the graph model, impeding inference for them. For instance, a drug-like chemical compound lacking significant structural similarity to existing drugs and without known protein targets becomes an isolated node in a drug-gene-disease graph. It becomes impractical and unreliable to infer its associations with diseases. 

Various machine learning-based automatic processes have been developed to enhance graph models, such as predicting gene-disease associations through Natural Language Processing\cite{multidcp_wu2022deep,cui2024scgpt, saturn_rosen2024toward,yang2023genecompass}, and drug-target interaction predictions\cite{portalcg_cai2023end,  chen2020transformercpi, li2022bacpi}. However, these predicted relationships may be inaccurate, resulting the introduction of false positives and conflicting relationships. Few attention have been paid to addressing the issue of dubious relationships in the knowledge graph, especially when it is generated from biomedical publications many of which cannot be reproduced\cite{prinz2011believe, begley2012raise, brito2020recommendations}. 

\section{AI-powered knowledge-enriched multi-scale genotype-environment-phenotype predictive modeling}
Recent advances in deep learning, coupled with the growing accessibility of multi-omics data, have opened avenues for predicting emergent phenotypes through novel perturbations under diverse genotypes. Leveraging these developments, we propose two complementary approaches and their combinations: (1) biology-inspired end-to-end multi-modal multi-task deep learning, (2)physics-informed context-specific multi-scale knowledge graphs.

\subsection{Biology-inspired end-to-end multi-modal multi-task deep learning 
}
\begin{figure}[h]
    \hspace{-4cm}
    \includegraphics[width=1.5\textwidth]{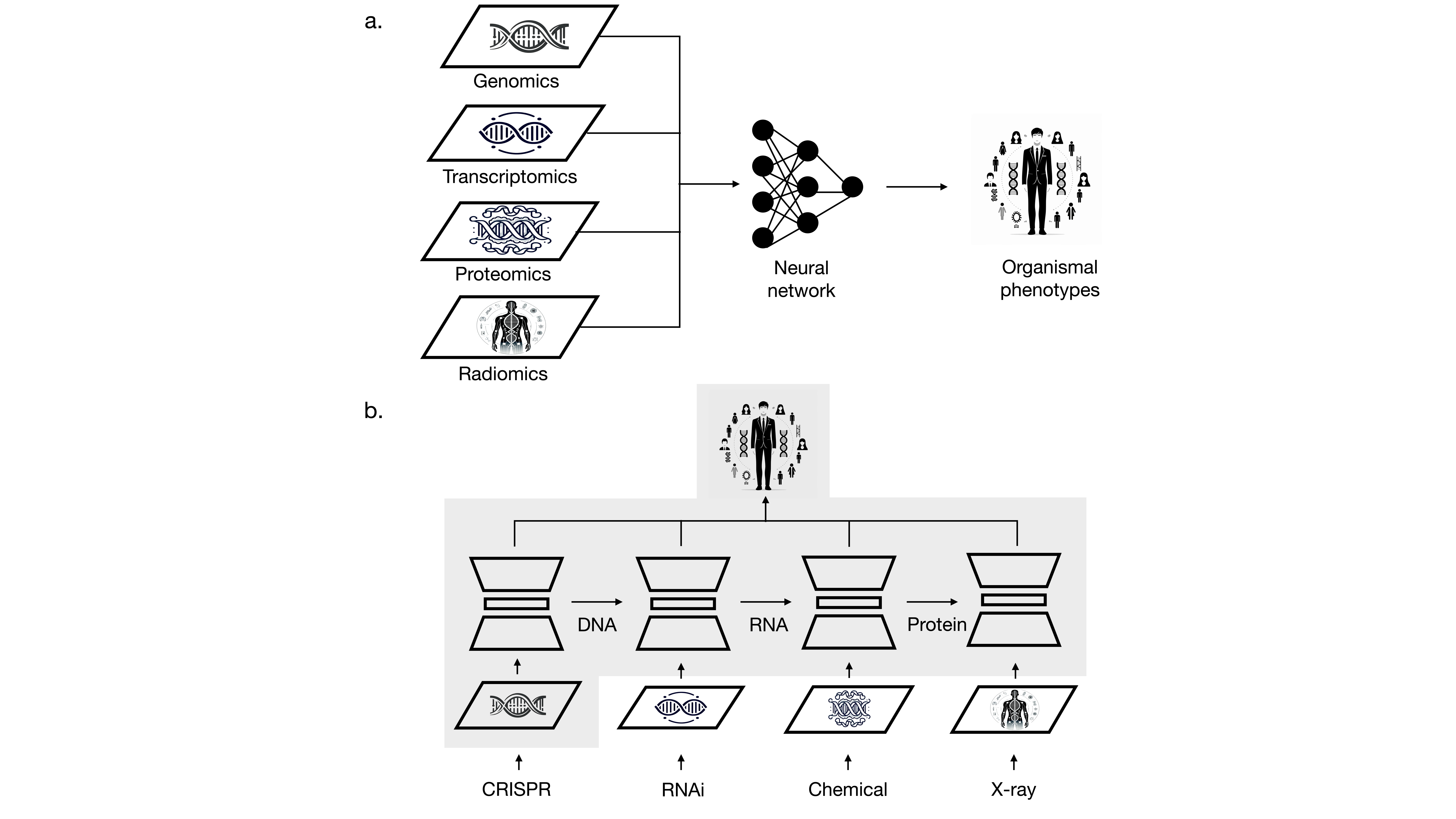}
    \caption{Illustration of multi-modal supervised learning. (a) A conventional strategy that requires paired data for all the modalities simultaneously. (b) An end-to-end deep neural network explicitly models asymmetric information flows from DNAs to RNAs to proteins to metabolites and ultimately to the organismal phenotype. Each modality can be pre-trained using unlabeled data. The paired data is used to fine-tune the model between any two modalities. After the model is fully trained, phenotypes can be predicted from genotypes through endophenotypes even if their data is not available.}
    \label{fig:our_flow}
\end{figure}

Compared to classical machine learning, one of the unique features of deep neural networks is their capacity for end-to-end learning. End-to-end learning tackles a complex task from inception to completion, as opposed to dividing the task into smaller sub-tasks and addressing them independently. In the context of predictive modeling for genotype-environment-phenotype relationships, an end-to-end deep neural network explicitly models asymmetric information flows from DNAs to RNAs to proteins to metabolites and ultimately to the organismal phenotype, following the central dogma of molecular biology, as illustrated in Figure \ref{fig:our_flow}. A foundation model for each data modality can be pre-trained and fine-tuned using modality-specific unlabeled and labeled data. When paired data across two biological levels is available, the models from different levels can be connected through contrastive learning\cite{he2022cross}, transfer learning\cite{transpro_wu2023hierarchical}, or other techniques\cite{GuidedSTab_wu2023multitask}. With labeled organismal phenotype data, all modalities are interconnected and fine-tuned from genotypes to phenotypes. Environmental factors can be applied to any level, contingent on the nature of influences and perturbations — examples include CRISPR-Cas9 on DNA, RNAi on RNA, and small molecule inhibitors on proteins. Utilizing a fully-trained end-to-end model, it becomes feasible to incorporate endophenotype information, even if it cannot be directly obtained (such as brain tissue proteomics for a living AD patient), thereby improving predictions of organismal phenotypes from a genotype.

\newpage

The biology-inspired end-to-end model can address the OOD and label scarcity problem from various perspectives. The pre-trained foundation model has exhibited notable generalization capabilities. For instance, the protein language model has proven successful in tasks such as protein structure predictions\cite{alphafold_jumper2021highly}, protein design\cite{madani2023large}, and predicting protein-chemical interactions\cite{portalcg_cai2023end}. Contrastive learning has proven successful in integrating multi-omics data, as demonstrated in the previous section.  Notably, several proof-of-concept studies have shown the promise of end-to-end models that adhere to the multi-level organization of a biological system.  For example, the Cross-Level Information Transmission (CLEIT) network employs transcriptomics endophenotypes as an intermediate layer to connect genomic mutations with cellular phenotypes through contrastive learning\cite{he2022cross}. This approach enhances phenotype predictions from genotypic data. Leveraging transfer learning, TransPro predicts proteomics profiles induced by unobserved chemicals based on transcriptomics data\cite{transpro_wu2023hierarchical}. It is observed that predicting organismal phenotypes via predicted and imputed proteomics signatures by TransPro is more accurate than relying on experimentally determined transcriptomics or proteomics data, which often suffer from noise and sparsity. Combining contrastive learning with multi-task learning guided by clinical features, Guided-Stab achieved survival prediction by cancer transcriptomics\cite{GuidedSTab_wu2023multitask}. An end-to-end model, which links genotypes to phenotypes by integrating multiple endophenotypes based on their biological relationships, is anticipated to offer a robust tool for establishing causal genotype-environment-phenotype relationships.

\subsection{Personalized physics-informed multi-scale knowledge graph}
Considering the elevated incidence of false negatives and false positives in relationships, as well as the presence of coarse-grained and ambiguous phenotypes in current biological network models, we propose three solutions to harness the potential of graph learning for predictive modeling of genotype-environment-phenotype relationships. These solutions comprise (1) the explicit representation of physical interactions within molecular networks, (2) the construction of context-dependent networks with fine-grained phenotypes, and (3) the development of multi-scale network models.

\begin{figure}[h]
    \hspace{-5cm}
    \includegraphics[width=1.5\textwidth]{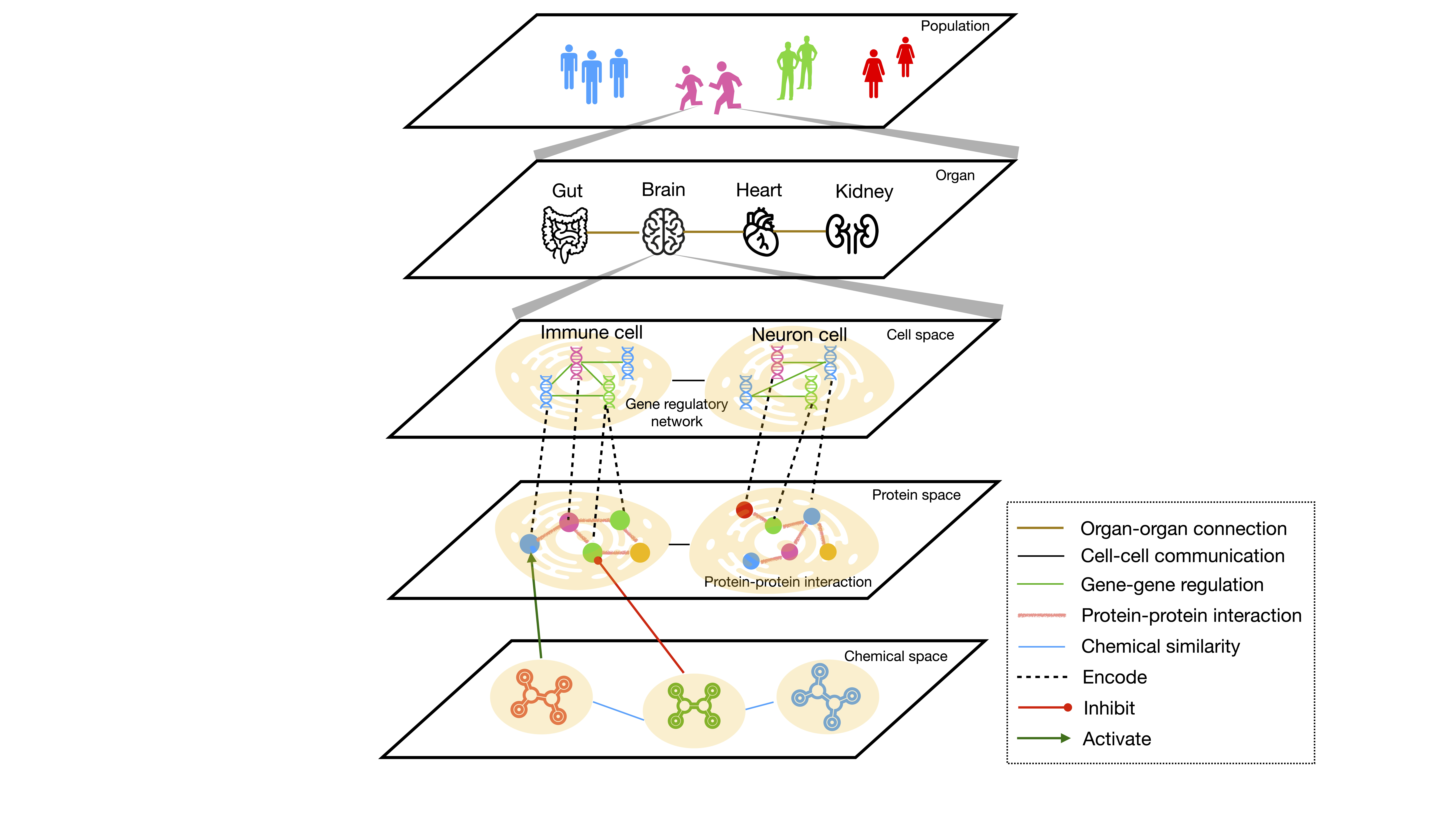}
    \caption{Illustration of personalized physics-informed multi-scale knowledge graph. It represents causal genotype-environment-phenotype from a single cell to an individual.}
    \label{fig:network}
\end{figure}

Genotype-phenotype relationships in many existing network models, such as gene-disease networks, primarily rely on statistical correlations derived from Genome-Wide Association Studies (GWAS). Without insight into the underlying molecular interactions, determining the molecular drivers responsible for a phenotype and predicting phenotypic responses to novel perturbations becomes challenging. By incorporating quantitative details of molecular interactions into the network, it becomes possible to rationalize how molecular changes may impact phenotypes. For example, mutations in DNA sequence can alter regulatory DNA-protein, regulatory RNA-protein, or protein-protein interactions, subsequently influencing the binding affinity or kinetics of these interactions, leading to changes in gene expression, signaling transduction, or metabolism. Illustrated in Figure \ref{fig:network}, representing these interactions in a network model with weighted and signed edges encoding the degree (or certainty) and direction of interaction changes allows for more confident inference of causal genotype-phenotype relationships\cite{liu2017varifunnet}. High-throughput techniques have emerged to explore understudied molecular interactions\cite{reinecke2023chemical,nechay2020high}.
New machine learning methods, e.g., model-agnostic semi-supervised meta-learning, can efficiently explore OOD drug-target interactions, metabolite-enzyme interactions, and microbiome metabolite-human receptor interactions\cite{mmaple}. Transfer learning enables predicting functional activities of ligand binding, i.e., antagonist vs agonist\cite{cai2022deepreal}.
\newpage

Many existing network models are canonical aggregations across different conditions. For instance, in a gene-disease network, "Alzheimer’s disease" (AD) is often depicted by a single node, and the gene-gene interaction network remains constant across all diseases. However, AD has several subtypes resulting from different etiologies (e.g., APOE4 vs. TREM2). Similarly, the gene-gene interaction network undergoes rewiring dependent on biological contexts (such as cell types, disease stages, and species). This coarse-grained representation falls short of capturing the complexities of biology. We propose to decompose the aggregated network model into an interconnected multiplex network model. Each plex in the network represents a subtype or an individual. In the case of a gene-disease network, using disentangled embeddings of disease biomarkers (e.g., brain imaging for AD), a subtype of AD or an individual patient (i.e., phenotype) can be represented by a class-specific embedding and a subtype/individual-specific embedding, which can be derived from patient-level data like medical imaging and electronic health records. Subtype/individual-specific gene-gene interaction networks can be derived from gene embeddings learned from a large language model\cite{cui2024scgpt,geneformer_theodoris2023transfer}. It is anticipated that such a fine-grained network model will be more potent in predictive modeling of genotype-environment-phenotype relationships compared to a coarse-grained aggregated model.

The inherent complexity and hierarchical organization of a biological system naturally lend themselves to representation on a multi-scale. For instance, a tissue can be portrayed through a cell-cell interacting network, and each cell can be captured by a cell type-specific gene-gene interacting network. Algorithmically, a multi-scale cell-cell interacting network can be conceptualized as a network of networks. While the network of networks concept has found widespread application in modeling areas such as the internet, smart cities, social networks, supply chains, telecommunications, cloud computing, and financial systems\cite{kenett2015networks}, its utilization in systems biology remains relatively limited\cite{schuster2006network}. Given the abundance of single-cell and spatial omics data, there is a compelling opportunity to explore the application of the network of networks paradigm for omics data integration and analysis in systems biology. 

\subsection{Integration of machine learning models, knowledge graphs, and generative AI}
The proposed machine learning and knowledge graph approaches mentioned above are complementary. Integrating these two approaches will further enhance the predictive power of genotype-environment-phenotype relationships. Although the machine learning model excels at discerning subtle patterns from raw data and augmenting missing links within a knowledge graph, it may lack a comprehensive understanding of the global contexts of these patterns. Conversely, a knowledge graph can consolidate patterns into a cohesive network within a broader context. Inference of missing links from a knowledge graph can both validate and refute predictions made by a machine learning model. 

\begin{figure}[h]
    \hspace{-3cm}
    \includegraphics[width=1.2\textwidth]{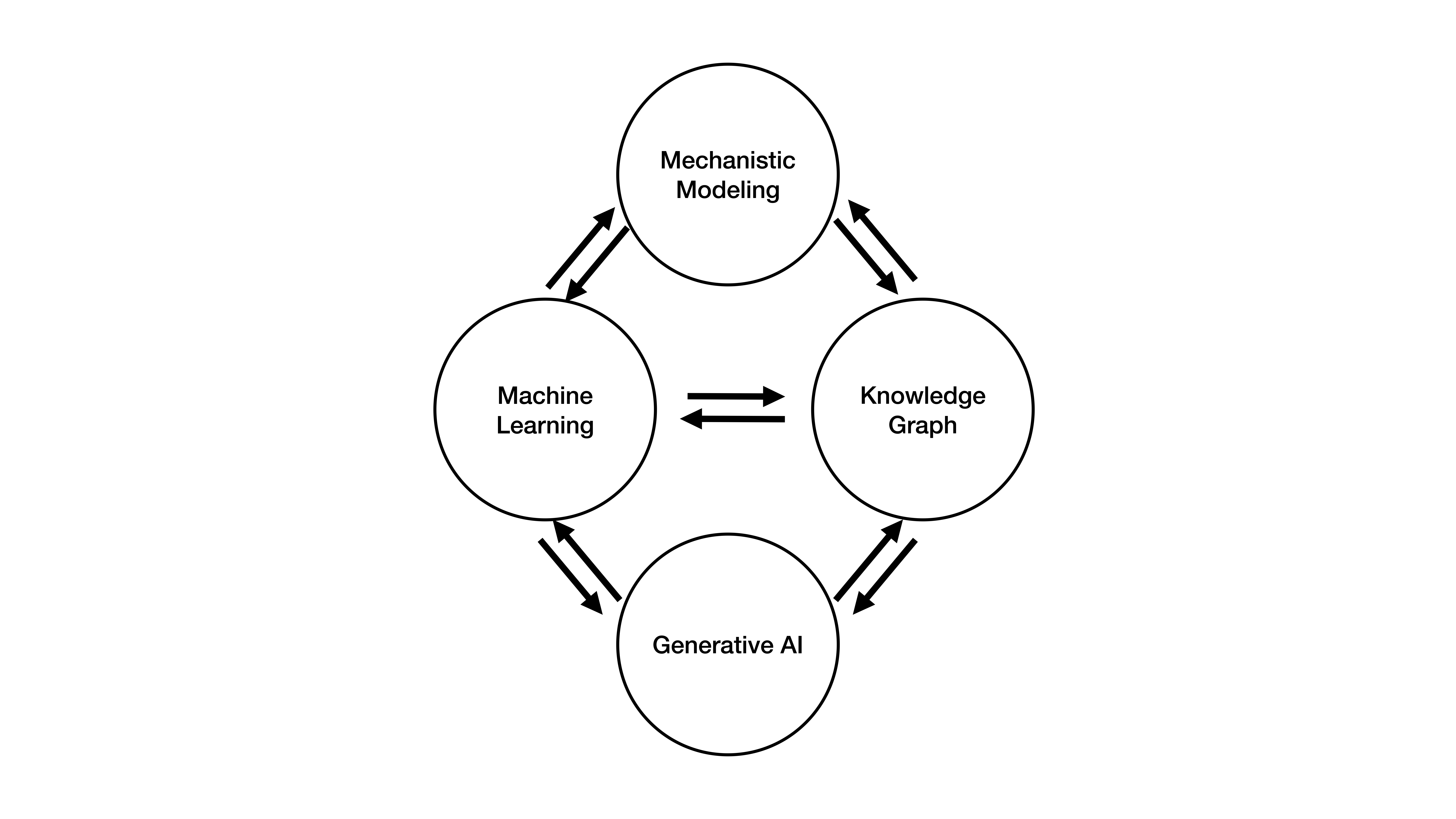}
    \caption{Integration of machine learning models, mechanistic models, knowledge graphs, and generative AI}
    \label{fig:ralation}
\end{figure}

Both machine learning models and knowledge graphs, which focus on predictive analytics, can benefit from integration with generative AI models. On one hand, a generative model can enhance predictive models in several ways. Generative models have the capability to generate synthetic data samples that closely resemble real data. These synthetic samples can effectively augment the training dataset of predictive models, particularly in scenarios where real data is limited. Furthermore, generative models can learn the underlying distribution of observed data, enabling them to identify outliers or OOD cases effectively. Additionally, they can be utilized to impute missing values by generating plausible values conditioned on the observed data. On the other hand, machine learning models can enhance personalization and mitigate hallucination in generative models through techniques such as reinforcement learning, attention mechanisms, conditional generation, active learning, and others \ref{fig:ralation}.

\section{Conclusion}
The fusion of multi-omics data and AI techniques marks a significant advancement in comprehending complex biological systems and predicting outcomes across diverse environments and perturbations. In this paper, we have explored the interleaved interactions between genotype, environment, and phenotype, highlighting the pivotal role of endophenotypes as intermediate markers linking genetic makeup to observable traits. Central to our discussion is the integration of multi-omics data, spanning various biological levels from single cells to whole organisms, and encompassing different data modalities and species. We have addressed the shortcomings of current machine learning methods, particularly in accurately predicting causal relationships between genotype, environment, and phenotype. Our proposed framework, inspired by biology and driven by AI, aims to untangle the complexities of living organisms and lay the groundwork for personalized medicine. 

It is important to underscore that AI alone cannot accomplish our objectives. A comprehensive representation of human biology and physiology needs a digital twin that captures micro and macro dynamics of the human body and its interactions with the environment\cite{digital_tang2024roadmap, national2023foundational, katsoulakis2024digital}. This necessitates the integration of AI with mechanism-based modeling, a promising technique for addressing challenges in machine learning. For example, constraint-based metabolic network modeling can predict organismal phenotypes directly, such as growth rates under diverse conditions. Unlike "black box" machine learning models, mechanism-based models explicitly represent system processes and interactions, offering insights into underlying principles. Leveraging existing knowledge, they can make predictions even with limited data, exhibiting greater generalizability across scenarios. Their transparency facilitates interpretation and understanding of influencing factors, crucial for applications like biomedicine. Additionally, the seamless integration of prior knowledge enhances prediction accuracy and relevance. In conclusion, A biology-inspired AI model, coupled with mechanism-based modeling, holds considerable promise for advancing our understanding of genotype-environmental-phenotype relationships and informing critical decision-making.

\newpage
\bibliographystyle{ieeetr}
\bibliography{ref}

\end{document}